\def\year{2021}\relax
\documentclass[letterpaper]{article}
\usepackage{aaai21}
\usepackage{times}
\usepackage{helvet}
\usepackage{courier}
\usepackage[hyphens]{url}
\usepackage{graphicx}
\usepackage{natbib}
\usepackage{caption}
\usepackage{enumitem}
\usepackage{amssymb}
\usepackage{booktabs}
\usepackage{romannum}
\usepackage{amsmath}

\title{PMVOS: Pixel-Level Matching-Based Video Object Segmentation\vspace{4.5mm}}

\author{Suhwan Cho, Heansung Lee, Sungmin Woo, Sungjun Jang, Sangyoun Lee\\}

\affiliations{
	Yonsei University, Seoul, Korea\\
	\{chosuhwan, hslee2860, smw3250, jeu2250, syleee\}@yonsei.ac.kr\\}

\begin{document}
\maketitle

\begin{abstract}
	Semi-supervised video object segmentation (VOS) aims to segment arbitrary target objects in video when the ground truth segmentation mask of the initial frame is provided. Due to this limitation of using prior knowledge about the target object, feature matching, which compares template features representing the target object with input features, is an essential step. Recently, pixel-level matching (PM), which matches every pixel in template features and input features, has been widely used for feature matching because of its high performance. However, despite its effectiveness, the information used to build the template features is limited to the initial and previous frames. We address this issue by proposing a novel method-PM-based video object segmentation (PMVOS)-that constructs strong template features containing the information of all past frames. Furthermore, we apply self-attention to the similarity maps generated from PM to capture global dependencies. On the DAVIS 2016 validation set, we achieve new state-of-the-art performance among real-time methods ($>$ 30 fps), with a $\mathcal{J}\&\mathcal{F}$ score of 85.6\%. Performance on the DAVIS 2017 and YouTube-VOS validation sets is also impressive, with $\mathcal{J}\&\mathcal{F}$ scores of 74.0\% and 68.2\%, respectively. 
\end{abstract}

\begin{figure}[t]
	\centering
	\includegraphics[width=1.0\linewidth]{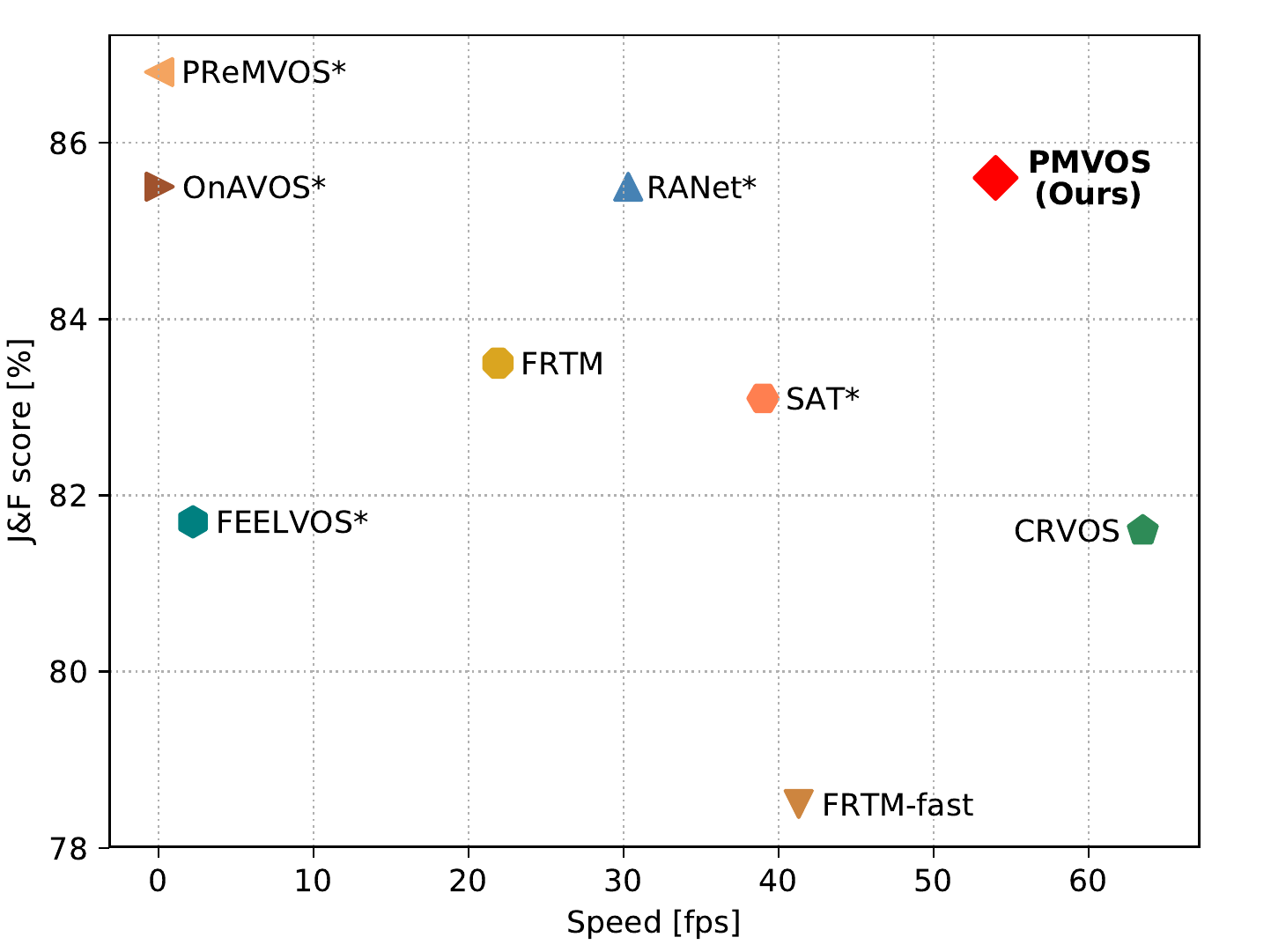}
	\caption{Comparison of state-of-the-art methods on the DAVIS 2016 validation set. We visualize the $\mathcal{J}$\&$\mathcal{F}$ score with respect to fps. $*$ indicates that the method uses a pre-trained segmentation model or synthetic training data. Our proposed PMVOS achieves the optimal speed-accuracy trade-off despite being trained on limited data.}
	\label{figure1}
\end{figure}

\section{Introduction}

\footnotetext{\vspace{2mm}}\makeatother

Video object segmentation (VOS) is an important task in computer vision because of its usefulness to diverse applications, including robotics, autonomous driving, video editing, and surveillance. This study focuses on a semi-supervised setting, which aims to track and segment arbitrary target objects in video. With a given ground truth segmentation mask of the initial frame, the goal is to segment a target object for all subsequent frames. Because the target object presented in the testing stage is arbitrary, it is not possible to train the network with information about the target object.

Previous studies, including OSVOS \cite{OSVOS}, OSVOS-S \cite{OSVOS-S}, OnAVOS \cite{OnAVOS}, and PReMVOS \cite{PReMVOS}, adapted the network to the arbitrary target object using an online learning strategy to train the network during the testing stage. Because this process enables a robust network for any type of target object, these methods demonstrate favorable accuracy on the benchmark datasets. However, because significant time is required during the testing stage, these methods are not practical.

Recent studies, including FEELVOS \cite{FEELVOS}, RANet \cite{RANet}, and A-GAME \cite{A-GAME}, overcome this significant time limitation by segmenting the target object without online learning. Instead, they focus on feature matching—they track and segment the target object strongly based on the comparison between template features representing the target object and input features. This approach is much more efficient than the online learning-based approach because the network can handle any type of arbitrary target object without time-consuming online network fine-tuning.

Pixel-level matching (PM), which matches every pixel in template features and input features, is a feature-matching technique that has received extensive attention due to its strong performance. Inspired by PM research, we propose a robust video object segmentation model using PM. We combine PM-based methods from previous studies to exploit their advantages and generate a strong baseline model.

Despite the high performance of PM-based methods, the information used to build template features is limited to the initial and previous frames. Therefore, we build a novel PM-based video object segmentation (PMVOS) model by constructing strong template features using the information about the target object from all past frames-thus overcoming the limited available information about the target object. Furthermore, self-attention is added to the multiple similarity maps generated from PM to reinforce information about the target object by capturing global dependencies.

Of the real-time methods with speeds greater than 30 fps, we achieve new state-of-the-art performance on the DAVIS 2016 \cite{DAVIS2016} validation set, with a $\mathcal{J}\&\mathcal{F}$ score of 85.6\%. Results on the DAVIS 2017 \cite{DAVIS2017} and YouTube-VOS \cite{YTVOS} validation sets are also competitive, with $\mathcal{J}\&\mathcal{F}$ scores of 74.0\% and 68.2\%, respectively. On these three popular benchmark datasets, we achieve a superior speed-accuracy trade-off-no other method has higher accuracy than the proposed method while maintaining a similar speed.

This study contributes four findings to existing research:
\begin{itemize}[leftmargin=0.2in]
	\item We propose a strong baseline model based on early PM-based methods that is more efficient than those methods. 
	\item We propose a novel method of building strong templates features for PM using all available information about the target object from the past frames-enabling the network to fully exploit appearance information about the target object. 	
	\item We apply self-attention to the multiple similarity maps generated from PM to reinforce information about the target object by capturing global dependencies.
	\item Experiments demonstrate that our proposed PMVOS achieves superior speed-accuracy trade-off on three popular benchmark datasets.
\end{itemize}

\section{Related Work}
\begin{figure*}[t]
	\centering
	\includegraphics[width=1.0\linewidth]{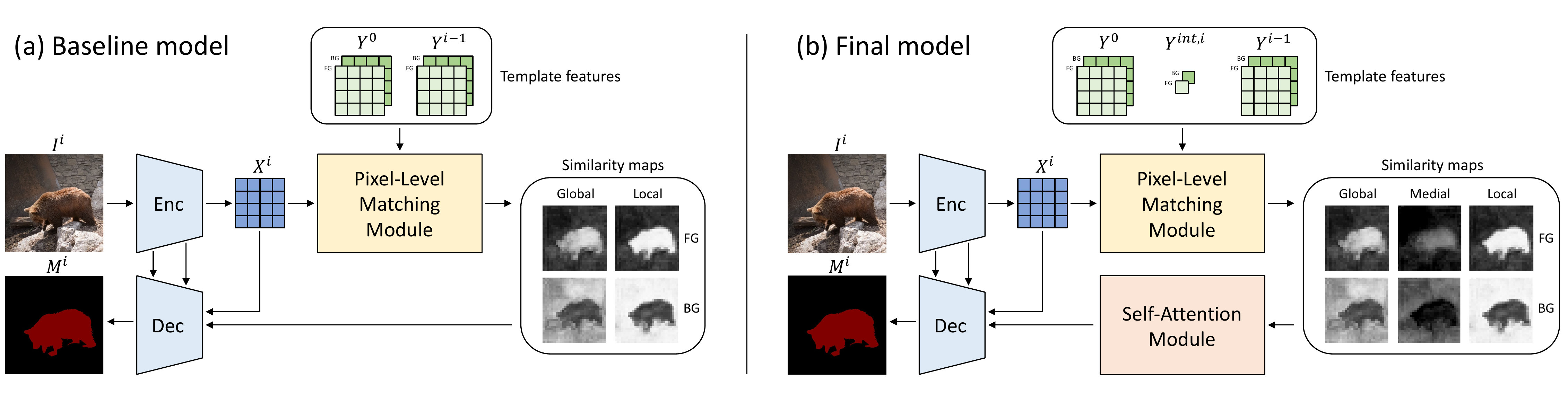}
	\caption{Illustration of our baseline and final PMVOS models. a) Our proposed baseline model is based on the early PM-based methods. Template features only contain the information of the initial and previous frames. b) Our final PMVOS model is developed based on our baseline model. Template features contain information from all past frames. Self-attention is applied to the similarity maps generated from PM to reinforce information about the target object by capturing global dependencies.}
	\label{figure2}
\end{figure*}

\vspace{1mm}
\noindent\textbf{Online learning:} Due to the difficulty of considering an arbitrary target object, early methods relied heavily on online learning. OSVOS \cite{OSVOS} trains the pre-trained foreground segmenting network during the testing stage with the ground truth segmentation mask given in the initial frame. Starting from OSVOS, OSVOS-S \cite{OSVOS-S} uses semantic information from an instance segmentation network, and OnAVOS \cite{OnAVOS} uses an online adaptation mechanism. DyeNet \cite{DyeNet} combines temporal propagation and a re-identification module and uses online learning to boost performance. PReMVOS \cite{PReMVOS} addresses the problem by separating it into proposal-generation, refinement, and merging with extensive online learning. Although the methods using online learning are more robust for arbitrary target objects than those without online learning, their practical use in real-world applications is limited because they are very time-consuming.

\vspace{1mm}
\noindent\textbf{Feature matching:} For practical usability, most recent studies focus on feature matching that compares template features and input features without using online learning. Feature matching is less effective than online learning but significantly more efficient. RGMP \cite{RGMP} uses a Siamese encoder and Global convolution block to compare the current frame’s features with the initial frame’s features. OSMN \cite{OSMN} uses two modulators
to capture instance-level information about the target object. A-GAME \cite{A-GAME} generates the appearance model of the target object using a Gaussian distribution and uses it as template features. TVOS \cite{TVOS} takes a label propagation approach based on feature similarity in an embedding space. The tracking-based methods are also in this category because bounding box generation is performed using feature matching. FAVOS \cite{FAVOS} uses a part-based tracking method. SiamMask \cite{SiamMask} performs both object tracking and VOS with a single approach, and SAT \cite{SAT} fuses object tracking and VOS into a truly unified pipeline.

\vspace{1mm}
\noindent\textbf{Pixel-level matching (PM):} PM-based feature matching methods, matching every pixel in template features and in input features, have gained attention recently due to their high performance. The use of PM can be categorized into global matching or local matching. Global matching uses the initial frame’s features as template features and the current frame’s features as input features. In contrast, local matching uses the previous frame’s features as template features and the current frame’s features as input features. VideoMatch \cite{VideoMatch} produces foreground and background similarity maps using global matching, and the final prediction is generated by applying the softmax function to the similarity maps. FEELVOS \cite{FEELVOS} designs a simple, fast, and strong model using a pre-trained semantic segmentation model as its backbone with global matching and local matching. RANet \cite{RANet} uses global matching with a ranking attention module to rank and select conformable feature maps according to their importance. In contrast to previous PM-based methods, we also use the intermediate frames to build the template features to perform medial matching.

\vspace{1mm}
\noindent\textbf{Self-attention:} Self-attention was originally proposed by research \cite{attention} on machine translation to compute the context at each position as a weighted sum of all positions. Due to its superiority in capturing long-range dependencies, it is applicable to various image and video problems in computer vision. For semantic segmentation, which is related to our research, DANet \cite{DANet} proposes a position attention module and channel attention module to adaptively integrate local features with their global dependencies. CFNet \cite{CFNet} aggregates the co-occurrent contexts with their co-occurrent probabilities across the spatial locations in a self-attention manner. CCNet \cite{CCNet} applies a criss-cross attention module to every pixel to harvest the contextual information of all pixels on its criss-cross path. In this study, we use self-attention to similarity maps, which function as the extracted features in semantic segmentation models.

\section{Proposed Method}
In this section, we first describe the architectures of our baseline and final models. Then, we introduce the proposed PM and self-attention modules. Finally, we present the network training scheme and implementation details of our method.

\subsection{Network Overview}
We propose PMVOS for fast and robust semi-supervised VOS. We combine previous PM-based methods to exploit their advantages and generate our baseline model. The overall structure is similar to RANet \cite{RANet}, and the template features are generated using the same approach as FEELVOS \cite{FEELVOS}. For the matching operation, we use cosine similarity, like VideoMatch \cite{VideoMatch}. Moreover, information of the intermediate frames is used to construct template features, and self-attention is added to capture global dependencies in the similarity maps. Our proposed PMVOS consists of four parts: (1) an encoder to extract the features from the current frame’s image, (2) a PM module to generate cosine similarity maps by comparing template features and input features, (3) a self-attention module to reinforce information about the target object in the similarity maps, and (4) a decoder generating final segmentation by merging and refining all extracted information. Our baseline and final models are illustrated in Figure \ref{figure2}. The previous frame’s predicted segmentation mask is also supplied to the decoder for mask propagation.

\subsection{PM Module}
PM is a popular feature-matching technique in semi-supervised VOS. By comparing every pixel in template features and input features, a complete comparison can be performed. To date, PM has only been performed for global matching and local matching. Global matching uses the information of the initial frame to construct template features, and local matching uses the information of the previous frame to construct template features. In contrast to early methods, we also propose medial matching that uses the information of the intermediate frames for template features.

Denote $I^i \in [0, 1]^{3 \times H^0 \times W^0}$ and $X^i \in [0, 1]^{C \times H \times W}$ as the input image at frame $i$ and normalized features extracted from the input image at frame $i$, respectively. The predicted probabilities of background and foreground at frame $i$ are given by $M^i_0$ and $M^i_1$ respectively, where $M^i \in [0, 1]^{H^0 \times W^0}$. For calculations with $X$, a downsampled version of the predicted probabilities, denoted as $m^i \in [0, 1]^{H \times W}$, is also assigned for foreground and background. Then, we have
\begin{equation}
Y^i = \{Y^i_k \; | \; Y^i_k = X^i \odot m^i_k\}_{k \in \{0, 1\}}
\end{equation}
which indicates probability-weighted normalized features for each class. Because the sums of the probabilities of foreground and background are equal to one at every spatial location, the sums of $Y^i_0$ and $Y^i_1$ should be equal to $X^i$.

Template features for global matching and local matching can be easily determined to $Y^0$ and $Y^{i-1}$ because they only cover a single frame. However, for medial matching, which aims to exploit the information of the intermediate frames, using $Y$ of all intermediate frames as template features is expensive in terms of computation and memory. We assign a single pixel each for foreground and background to update the features instead of storing all of them, thus using the intermediate frames efficiently. We first define temporarily assigned template features, $\tilde{Y}^{int}$, at frame $i$ as
\begin{equation}
\tilde{Y}^{int, i} = \frac{\sum_p {_pY^i}} {\sum_p {_pm^i}}
\end{equation}
where $p$ covers every spatial location in $Y$ and $m$, i.e., covers $H^0 \times W^0$. Then, the template features for medial matching $Y^{int}$ at frame $i$ can be obtained by
\begin{equation}
{Y}^{int, i} = (1-\alpha)Y^{int, i-1} + \alpha\tilde{Y}^{int, i}
\end{equation}
where $\alpha$ is a learning rate for the feature update. At the initial frame, $\alpha$ is set to $1$ since there is no previous frame, and at the subsequent frames, $\alpha$ is set to $0.1$. The choice of $\alpha$ is based on empirically measuring the accuracy of the network.

\begin{figure}[t]
	\centering
	\includegraphics[width=1.0\linewidth]{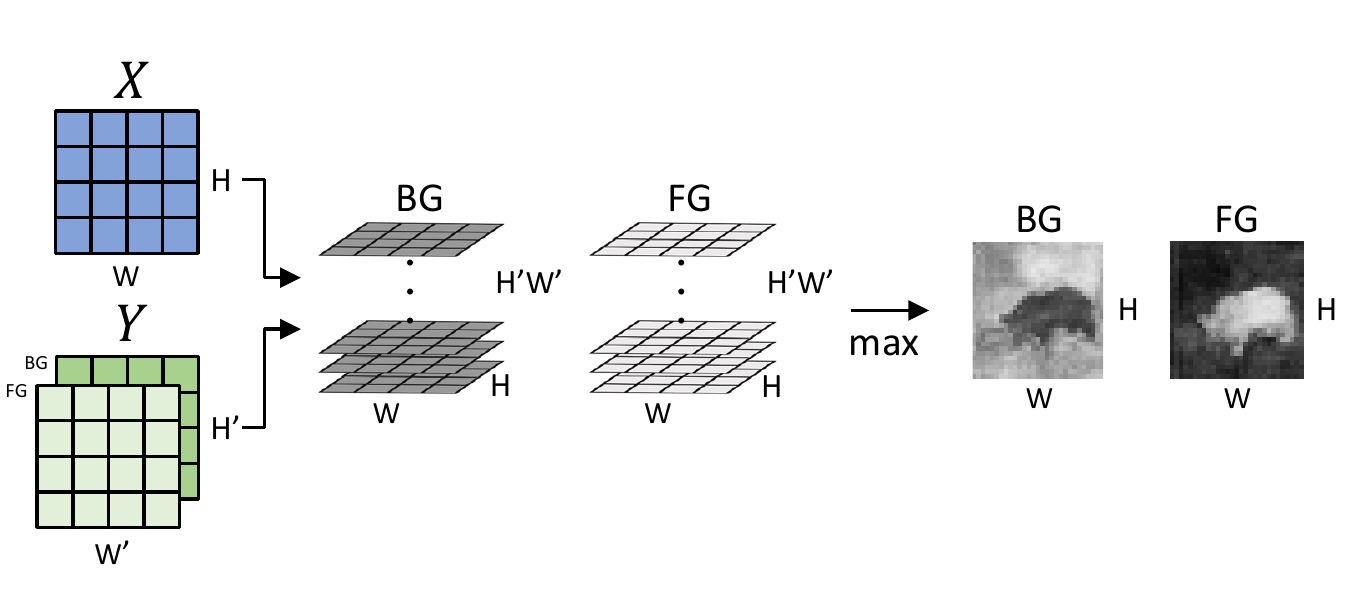}
	\caption{Mechanism of PM. For foreground and background, a cosine similarity vector with a size of $H'W' \times H \times W$ is generated for each, and by taking the maximum value in each channel, the final similarity map with a size of $H \times W$ is obtained.}
	\label{figure3}
\end{figure}

We can now perform PM by comparing template features and input features pixel-by-pixel. The mechanism of PM is visualized in Figure \ref{figure3}. The spatial sizes of $X$ and $Y$ are set to $H \times W$ and $H' \times W'$, respectively. By implementing convolution operations for every pixel in $X$ and every pixel in $Y$, probability-weighted cosine similarity vectors with a size of $H'W' \times H \times W$ are generated for foreground and background. The final similarity map with a size of $H \times W$ can be obtained by using the maximum value in each channel. In the baseline model, the PM is performed for global matching and local matching; therefore, two maps, one each for foreground and background, are generated. However, in the final model, the PM is performed for medial matching, global matching, and local matching, to fully use the information of all past frames, so three maps for each class are generated.

\subsection{Self-Attention Module}
Self-attention is an effective technique in computer vision, with its ability to capture global dependencies. It is also widely used in most state-of-the-art methods for semantic segmentation, which is similar to VOS. Therefore, we use self-attention for the similarity maps generated from PM, since the similarity maps in VOS function as high-level features in semantic segmentation.

We apply the attention mechanism to the similarity maps by first dividing the similarity maps into foreground and background maps. Then, for each bundle of maps, we use both spatial and channel attention to capture global dependencies in both spatial and channel dimensions, using a similar method as DANet \cite{DANet}. Spatial attention is used to capture a wider range of contextual relationships into local features, and channel attention is used to exploit the interdependencies between each channel (i.e., the similarity maps generated by different PM operations).

\begin{figure}[t]
	\centering
	\includegraphics[width=1.0\linewidth]{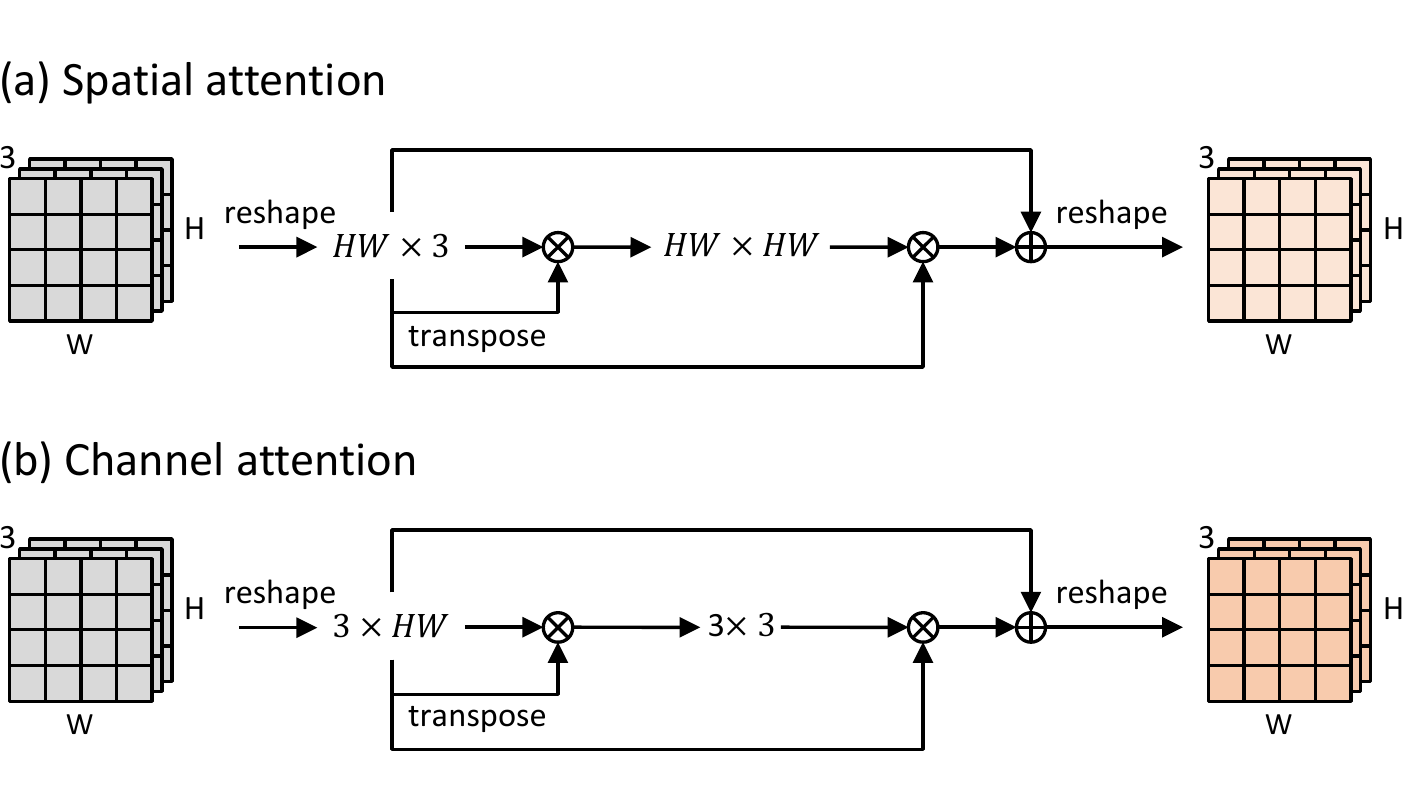}
	\caption{Mechanism of self-attention in our method. a) Spatial attention to capture spatial global dependencies. b) Channel attention to capture interdependencies between different similarity maps.}
	\label{figure4}
\end{figure}
In Figure \ref{figure4}, we visualize the mechanisms of spatial attention and channel attention, which are separately applied to foreground and background maps. For spatial attention, the input is reshaped into $HW \times 3$ to generate $HW \times HW$ features to capture spatial dependencies. For channel attention, the input is reshaped into $3 \times HW$ to generate $3 \times 3$ features, which are used to capture interdependencies between different similarity maps.  In the final model, the self-attention module uses a size of $3 \times H \times W$ maps for its input and outputs $6 \times H \times W$ features for each class. Then, the output is supplied to the decoder as important appearance information about the target object.

\subsection{Network Training}
We use a three-step training strategy. In the first stage, we train the network with a low resolution, $240 \times 432$, on both DAVIS 2017 \cite{DAVIS2017} and YouTube-VOS \cite{YTVOS} training sets for $80$ epochs. Each snippet is composed of $8$ frames with a batch size of $4$. In the second stage, the network is trained with a size of $480 \times 864$ on the same datasets for $50$ epochs. Each snippet is composed of $12$ frames with a batch size of $2$ to capture the long-term variations of the target object. In the last stage, the network is fine-tuned on the DAVIS 2017 training set with the original resolution for $80$ epochs, and each snippet has the same form as in the second stage. 
If there are multiple target objects, we randomly select one in each batch. For all stages, a cross-entropy loss is applied, and the Adam \cite{Adam} optimizer is used to minimize the loss. The learning rate is set to $10^{-4}$, $10^{-5}$, and $10^{-5}$, with decays in the exponential learning rate of $0.975$, $0.985$, and $1.0$ per epoch for the first, second, and last stages, respectively. For data augmentation, only a horizontal flip is used. Full training requires $32$ hours on a single GeForce GTX Titan X GPU.

\subsection{Implementation Details}
We use ResNet-50 \cite{resnet} pre-trained on ImageNet \cite{imagenet} as our encoder, and the features extracted from the last three blocks are used to capture the multi-scale information. Each block has a spatial size of $1/4$, $1/8$, and $1/16$ of the input size, and a channel size of $256$, $512$, and $2048$, respectively. For the decoder, we follow the method of CRVOS \cite{CRVOS}. The decoder refines and merges the input features gradually with multi-scale features extracted from the encoder. A deconvolution layer is used to upscale the features, instead of a bilinear upsampling layer, to generate a spatially robust segmentation mask.

\section{Experiments}
In this section, we first describe the datasets and evaluation metrics used in this study. We validate the effectiveness of each component in the network by performing both quantitative and qualitative ablation studies. The results of our method are then compared with other state-of-the-art methods on three popular benchmark datasets. Finally, we visualize and discuss the resulting segmentation masks from our method.
\begin{table}[t]
	\centering
	\begin{tabular}{l | c c c c | c c c}
		&G &L &M &A &D16 &D17 &YT\\ \midrule
		\Romannum{1} &\checkmark & & & &81.4 &66.9 &62.9\\
		\Romannum{2} (Baseline) &\checkmark &\checkmark & & &84.9 &69.3 &64.7\\
		\Romannum{3} &\checkmark &\checkmark &\checkmark & &85.0 &72.6 &66.7\\
		\Romannum{4} (Final) &\checkmark &\checkmark &\checkmark &\checkmark &\textbf{85.6} &\textbf{74.0} &\textbf{68.2}\\
	\end{tabular}
	\caption{Ablation study for each component of our proposed PMVOS. G, L, M, and A denote global matching, local matching, medial matching, and self-attention, respectively. D16, D17, and YT indicate $\mathcal{J}\&\mathcal{F}$ scores on DAVIS 2016, DAVIS 2017, and YouTube-VOS, respectively.}
	\label{table1}
\end{table}

\subsection{Experimental Setup}
\vspace{1mm}
\noindent\textbf{Datasets:} Semi-supervised VOS is associated with three popular benchmark datasets. DAVIS 2016 \cite{DAVIS2016} comprises a total of 50 video sequences-30 for the training set and 20 for the validation set. Each sequence contains a single target object. DAVIS 2017 \cite{DAVIS2017} is an extended version of DAVIS 2016, with training and validation sets composed of 60 and 30 video sequences, respectively, and 2.3 and 1.97 mean target objects, respectively. YouTube-VOS \cite{YTVOS} is the largest dataset for semi-supervised VOS, containing 3,471 video sequences in the training set and 507 in the validation set. The videos are 24 fps in DAVIS 2016 and DAVIS 2017, and 30 fps in YouTube-VOS. In DAVIS 2016 and DAVIS 2017, every frame is annotated, but in YouTube-VOS, every five frames is annotated.

\vspace{1mm}
\noindent\textbf{Evaluation Metrics:} The predicted segmentation masks are evaluated using region accuracy and contour accuracy. Region accuracy, $\mathcal{J}$, can be obtained by computing the number of pixels of the intersection between the predicted segmentation mask and the ground truth segmentation mask and then dividing it by the size of the union. For contour accuracy, we first perform bipartite matching between the boundary pixels of the predicted segmentation mask and the ground truth segmentation mask. Then, contour accuracy, F, can be obtained by calculating the $\mathcal{F}$ measure between the precision and recall of the matching.

\subsection{Ablation Study}
Table \ref{table1} presents the quantitative ablation studies for each component of our method on three popular benchmark datasets. \Romannum{1} is the model that only uses global matching for feature matching, and \Romannum{2} uses both global matching and local matching. Accuracy is boosted 2-3\% on all benchmark datasets when using local matching additionally. \Romannum{2} is our baseline model, which is based on the early PM methods. \Romannum{3} performs global matching, local matching, and medial matching for PM to fully exploit the information from all past frames. Using intermediate frames with medial matching is effective, with a 3.3\% accuracy improvement on the DAVIS 2017 \cite{DAVIS2017} validation set, and 2\% on the YouTube-VOS \cite{YTVOS} validation set. \Romannum{4} is our final model, with global matching, local matching, medial matching, and self-attention for the similarity maps. Adding self-attention produces an approximately 1\% accuracy improvement for the network on all benchmark datasets.

\begin{figure}[t]
	\centering
	\includegraphics[width=1.0\linewidth]{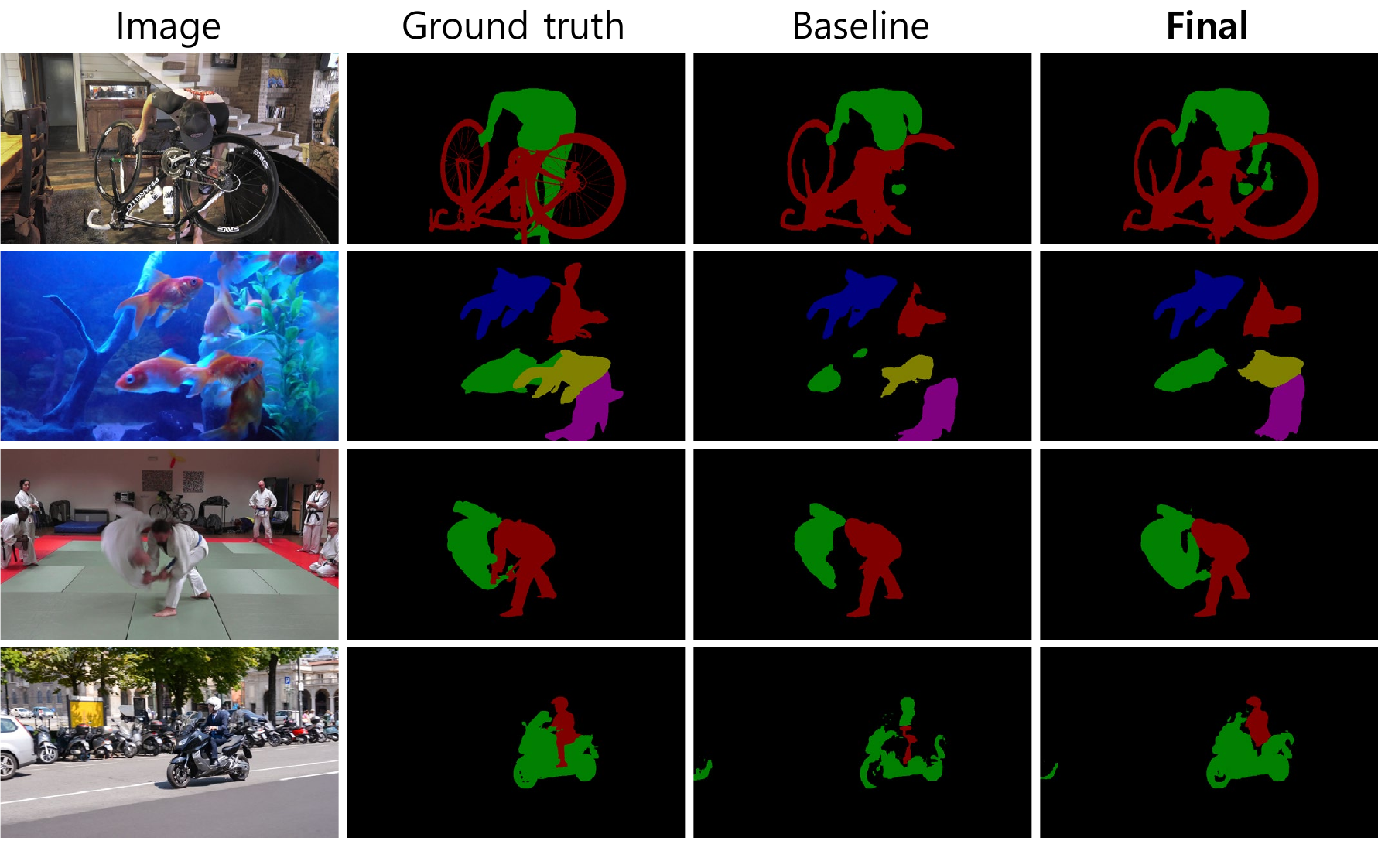}
	\caption{Qualitative comparison of our baseline and final models on DAVIS 2017 validation set. Columns, from left to right: input image, ground truth segmentation mask, predicted segmentation mask of our baseline model, and our final model.}
	\label{figure5}
\end{figure}

We also present a qualitative comparison of our baseline and final models in Figure \ref{figure5}. There is no extreme difference between their result segmentation masks; however, the final model generates more accurate details compared to the baseline model. This is because template features in the final model have more robust information about the target object, and the final segmentation masks are generated knowing the global dependencies in the similarity maps. Both quantitative and qualitative results demonstrate the effectiveness of each component.

\begin{table*}[t]
	\centering
	\begin{tabular}{l| c c | c | c c c | c c c} \toprule
		Method &seg &syn &fps &$\mathcal{J}\&\mathcal{F}_{16}$ &$\mathcal{J}_{16}$ &$\mathcal{F}_{16}$ &$\mathcal{J}\&\mathcal{F}_{17}$ &$\mathcal{J}_{17}$ &$\mathcal{F}_{17}$\\ \midrule
		PReMVOS (\citeauthor{PReMVOS}) &\checkmark &\checkmark &0.03 &\textbf{86.8} &84.9 &\textbf{88.6} &\textbf{77.8} &\textbf{73.9} &\textbf{81.7}\\
		OnAVOS (\citeauthor{OnAVOS}) &\checkmark & &0.08 &85.5 &\textbf{86.1} &84.9 &67.9 &64.5 &71.2\\
		OSVOS (\citeauthor{OSVOS}) &\checkmark & &0.11 &80.2 &79.8 &80.6 &60.3 &56.6 &63.9\\
		OSVOS-S (\citeauthor{OSVOS-S}) &\checkmark & &0.22 &86.5 &85.6 &87.5 &68.0 &64.7 &71.3\\
		FAVOS (\citeauthor{FAVOS}) & & &0.56 &81.0 &82.4 &79.5 &58.2 &54.6 &61.8\\
		FEELVOS (\citeauthor{FEELVOS}) &\checkmark & &2.22 &81.7 &81.1 &82.2 &71.5 &69.1 &74.0\\
		VideoMatch (\citeauthor{VideoMatch}) &\checkmark & &3.13 &80.9 &81.0 &80.8 &62.4 &56.5 &68.2\\ 
		OSMN (\citeauthor{OSMN}) & &\checkmark &7.14 &73.5 &74.0 &72.9 &54.8 &52.5 &57.1\\
		RGMP (\citeauthor{RGMP}) & &\checkmark &7.69 &81.8 &81.5 &82.0 &66.7 &64.8 &68.6\\
		FTMU (\citeauthor{FTMU}) &\checkmark & &11.1 &78.9 &77.5 &80.3 &70.6 &69.1 &72.1\\
		DMM-Net (\citeauthor{DMM-Net}) &\checkmark & &12.0 &- &- &- &70.7 &68.1 &73.3\\
		A-GAME (\citeauthor{A-GAME}) & &\checkmark &14.3 &82.1 &82.0 &82.2 &70.0 &67.2 &72.7\\
		FRTM (\citeauthor{FRTM}) & & &21.9 &83.5 &- &- &76.7 &- &-\\ \midrule
		RANet (\citeauthor{RANet}) & &\checkmark &30.3 &85.5 &85.5 &\textbf{85.4} &65.7 &63.2 &68.2\\
		TVOS (\citeauthor{TVOS}) & & &37.0 &- &- &- &72.3 &69.9 &74.7\\ 
		SAT (\citeauthor{SAT}) &\checkmark & &39.0 &83.1 &82.6 &83.6 &72.3 &68.6 &76.0\\
		FRTM-fast (\citeauthor{FRTM}) & & &41.3 &78.5 &- &- &70.2 &- &-\\
		\textbf{PMVOS (Ours)} & & &54.0 &\textbf{85.6} &\textbf{86.1} &85.1 &\textbf{74.0} &\textbf{71.2} &\textbf{76.7}\\
		SiamMask (\citeauthor{SiamMask}) &\checkmark & &55.0 &69.8 &71.7 &67.8 &56.4 &54.3 &58.5\\
		SAT-fast (\citeauthor{SAT}) &\checkmark & &60.0 &- &- &- &69.5 &65.4 &73.6\\
		CRVOS (\citeauthor{CRVOS}) & & &63.5 &81.6 &82.2 &81.0 &54.3 &53.5 &55.1\\ \bottomrule
	\end{tabular}
	\caption{Quantitative evaluation on DAVIS 2016 and DAVIS 2017 validation sets using fps and $\mathcal{J}\&\mathcal{F}$ score. DAVIS 2016 and DAVIS 2017 are denoted by 16 and 17, respectively. The methods using a pre-trained segmentation model and synthetic training data are marked seg and syn, respectively. The speed is calculated on the DAVIS 2016 validation set. If there is no publicly available speed, we use the speed of the DAVIS 2017 or YouTube-VOS validation sets.}
	\label{table2}
\end{table*}

\subsection{Quantitative Results}
We evaluate and compare our proposed PMVOS with state-of-the-art methods on the validation sets of three popular benchmark datasets: DAVIS 2016 \cite{DAVIS2016}, DAVIS 2017 \cite{DAVIS2016}, and YouTube-VOS \cite{YTVOS}. The comparison is performed based on the $\mathcal{J}\&\mathcal{F}$ score (indicating segmentation accuracy) and fps (indicating the inference speed of the network). The speed of our method is calculated on a single GeForce RTX 2080 Ti GPU. For a fair comparison, if the method uses a pre-trained segmentation model trained on image segmentation datasets (like MS-COCO \cite{COCO} or PASCAL-VOC \cite{PASCAL}) or uses synthetic data for network training, we mark it as seg or syn, respectively.

\vspace{1mm}
\noindent\textbf{DAVIS 2016:} In Table \ref{table2}, we compare our method with state-of-the-art methods on the DAVIS 2016 validation set. The highest accuracy is achieved by PReMVOS \cite{PReMVOS} with a $\mathcal{J}\&\mathcal{F}$ score of 86.8\%, but this accuracy is relatively low, considering the incredibly slow inference speed due to extensive fine-tuning on the initial frame. Of the previous methods, RANet \cite{RANet} and SAT \cite{SAT} exhibit the highest practical performance, with accuracies in the range of 83\% to 85\% and relatively fast inference speeds. Our proposed PMVOS surpasses these methods for both speed and accuracy, at 54.0 fps and a $\mathcal{J}\&\mathcal{F}$ score of 85.6\%. To the best of our knowledge, we achieve a new state-of-the-art speed-accuracy trade-off on the DAVIS 2016 validation set. Our method is even competitive with online learning-based methods, including PReMVOS, OnAVOS \cite{OnAVOS}, and OSVOS-S \cite{OSVOS-S}, while being significantly faster than them. This is attained by training the network on limited data, not by using a pre-trained segmentation model or synthetic training data.

\begin{table}[t]
	\centering
	\begin{tabular}{l | c} \toprule
		Method &$\mathcal{J}\&\mathcal{F}$\\ \midrule
		PReMVOS (\citeauthor{PReMVOS}) &66.9\\
		OnAVOS (\citeauthor{OnAVOS}) &55.2\\
		OSVOS (\citeauthor{OSVOS}) &58.8\\
		S2S (\citeauthor{YTVOS}) &57.6\\
		RGMP (\citeauthor{RGMP}) &53.8\\
		CapsuleVOS (\citeauthor{CapsuleVOS}) &62.3\\
		A-GAME (\citeauthor{A-GAME}) &66.0\\
		FRTM (\citeauthor{FRTM}) &\textbf{72.1}\\ 
		RVOS (\citeauthor{RVOS}) &56.8\\ \midrule 
		TVOS (\citeauthor{TVOS}) &67.8\\
		SAT (\citeauthor{SAT}) &63.6\\
		FRTM-fast (\citeauthor{FRTM}) &65.7\\
		\textbf{PMVOS (Ours)} &\textbf{68.2}\\
		SiamMask (\citeauthor{SiamMask}) &52.8\\ \bottomrule
	\end{tabular}
	\caption{Quantitative evaluation on the YouTube-VOS validation set. Methods are sorted in order of speed.}
	\label{table3}
\end{table}

\begin{figure*}[t]
	\centering
	\includegraphics[width=1.0\linewidth]{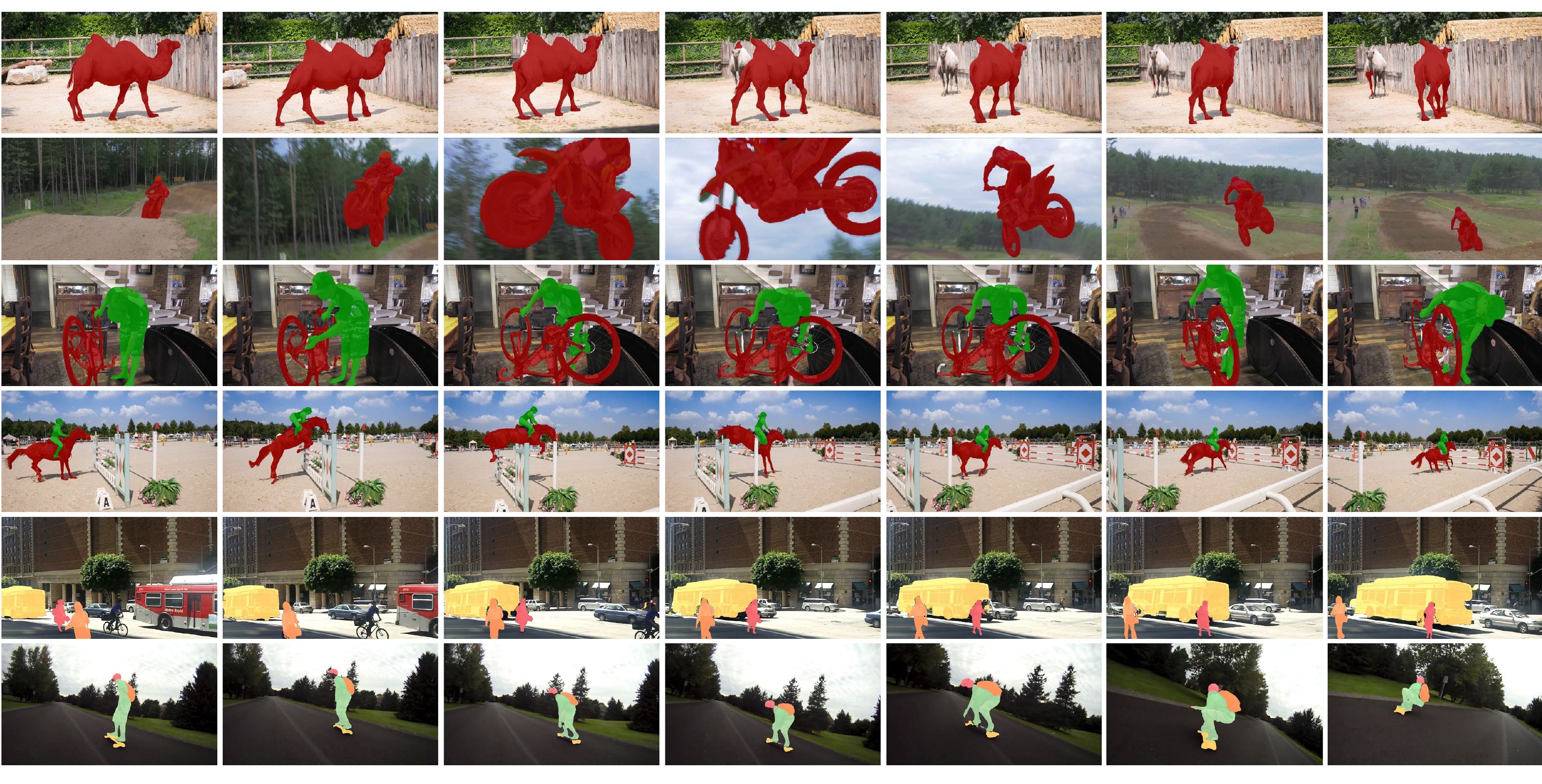}
	\caption{Qualitative results on the benchmark datasets. Two rows each in order from top to bottom are the results on DAVIS 2016, DAVIS 2017, and YouTube-VOS validation sets, respectively.}
	\label{figure6}
\end{figure*}
\begin{figure}[h]
	\centering
	\includegraphics[width=1.0\linewidth]{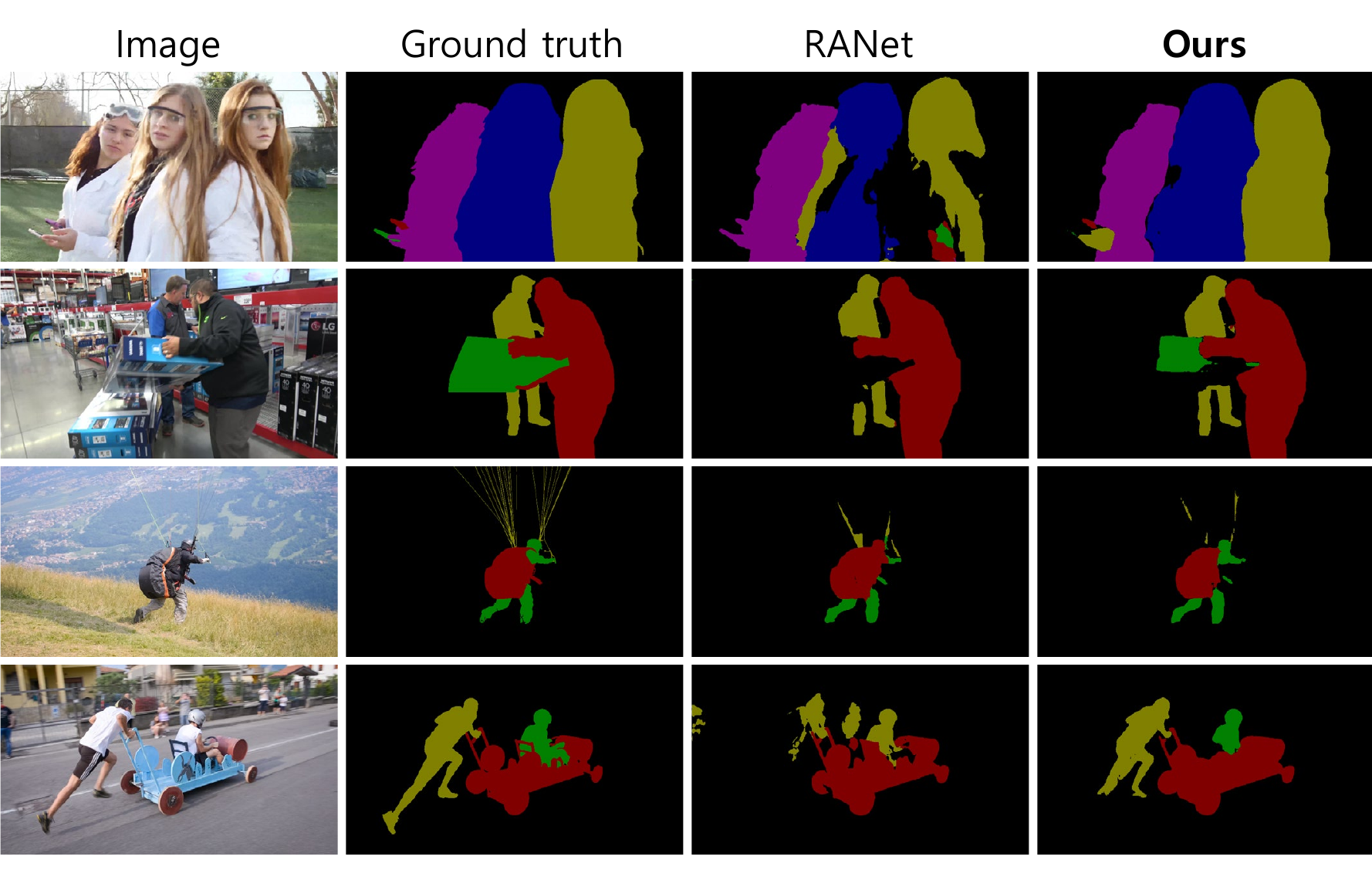}
	\caption{Qualitative comparison of RANet and the proposed method on DAVIS 2017 validation set. Columns from left to right: input image, ground truth segmentation mask, predicted segmentation mask of RANet, and the proposed PMVOS.}
	\label{figure7}
\end{figure}

\vspace{1mm}
\noindent\textbf{DAVIS 2017:} The results on the DAVIS 2017 validation set are also summarized in Table \ref{table2}. The highest accuracy is achieved by PReMVOS, with a $\mathcal{J}\&\mathcal{F}$ score of 77.8\%, but it is not impressive due to the slow inference speed compared to other recent methods. TVOS \cite{TVOS} and SAT achieve a significant speed-accuracy trade-off, with a $\mathcal{J}\&\mathcal{F}$ score of approximately 72\% and a real-time performance of 40 fps. However, the optimal speed-accuracy trade-off is achieved by our method, with a $\mathcal{J}\&\mathcal{F}$ score of 74.0\%. Among real-time methods running faster than 30 fps, we also achieve new state-of-the-art performance on the DAVIS 2017 validation set. Of course, our approach outperforms the early PM-based methods, including VideoMatch \cite{VideoMatch}, FEELVOS \cite{FEELVOS}, and RANet, by a large margin.

\vspace{1mm}
\noindent\textbf{Youtube-VOS:} In Table \ref{table3}, we compare our method with state-of-the-art methods on the YouTube-VOS validation set. FRTM \cite{FRTM} outperforms all other methods, achieving $\mathcal{J}\&\mathcal{F}$ score of 72.1\%, but its speed is not competitive compared to other recent methods with real-time performance. Of the real-time methods, TVOS and FRTM-fast have favorable accuracy; however, our method outperforms all other real-time methods, with a $\mathcal{J}\&\mathcal{F}$ score of 68.2\%. Our model is fine-tuned on the DAVIS 2017 training set, which indicates that the accuracy can be improved by separately fine-tuning the network on the YouTube-VOS training set.

\subsection{Qualitative Results}
In Figure \ref{figure6}, we illustrate the qualitative results of the proposed PMVOS on the DAVIS 2016 \cite{DAVIS2016}, DAVIS 2017 \cite{DAVIS2017}, and YouTube-VOS \cite{YTVOS} validation sets. Our method is robust for challenging scenarios, including fast appearance changes of the target object, various camera motions, distracting backgrounds, and occlusion between multiple target objects. We qualitatively compare the segmentation results with RANet \cite{RANet}, which is similar to our method, in Figure \ref{figure7}. Even though our method runs nearly twice as fast as RANet on the same device, the result segmentation mask is significantly more accurate than RANet.

\section{Conclusion}
We propose PMVOS, a strong semi-supervised VOS method based on PM. In contrast to the early PM-based methods, we fully use all past frames’ information to predict the current frame’s segmentation mask. Furthermore, we use multiple similarity maps by applying self-attention to capture global dependencies in the maps. Our method achieves a superior speed-accuracy trade-off on all benchmark datasets despite being trained on limited data. We believe that our method can be applied to fields in computer vision that require a strong ability to track and segment the target object with real-time performance.

\bibliography{references}

\end{document}